# Vision Mamba-based autonomous crack segmentation on concrete, asphalt, and masonry surfaces

Zhaohui Chen [a], Elyas Asadi Shamsabadi [a], Sheng Jiang [b], Luming Shen [a], Daniel Dias-da-Costa [a,*]

[a] *School of Civil Engineering, Faculty of Engineering, The University of Sydney, Sydney, NSW 2006, Australia*

[b] *College of Water Conservancy and Hydropower Engineering, Hohai University, Nanjing, 210098, China*

**Abstract:**

Convolutional neural networks (CNNs) and Transformers have shown advanced accuracy in crack detection under certain conditions. Yet, the fixed local attention can compromise the generalisation of CNNs, and the quadratic complexity of the global self-attention restricts the practical deployment of Transformers. Given the emergence of the new-generation architecture of Mamba, this paper proposes a Vision Mamba (VMamba)-based framework for crack segmentation on concrete, asphalt, and masonry surfaces, with high accuracy, generalisation, and less computational complexity. Having 15.6% - 74.5% fewer parameters, the encoder-decoder network integrated with VMamba could obtain up to 2.8% higher mDS than representative CNN-based models while showing about the same performance as Transformer-based models. Moreover, the VMamba-based encoder-decoder network could process high-resolution image input with up to 90.6% lower floating-point operations.

**Paper Highlights:**

- The first framework for crack segmentation with VMamba is proposed.
- The core select-scan module of VMamba has linear computational complexity.
- VMamba nearly matches Transformer performance with up to 74.5% fewer parameters.
- High-resolution input can be efficiently processed with VMamba-based architecture.

# 1. Introduction

A timely detection of defects in structural elements is crucial for ensuring the safety and longevity of infrastructure. The early identification of deficiencies can allow for an early intervention, thus reducing the likelihood of major failures. Nonetheless, the traditionally employed manual inspection methods have inherent drawbacks, including but not limited to being time-consuming, labour-intensive, and subject to human errors and biases [1,2]. Moreover, manual inspections can be hazardous, particularly when assessing hard-to-reach or unsafe areas. The subjective nature of human judgment in identifying defects may also lead to inconsistencies and inaccuracies in assessments [3,4]. As such, there is a growing need for automated and efficient defect detection techniques. Technologies such as computer vision, machine learning (ML), and remotely operated vehicles can streamline inspection and enhance infrastructure safety [5-7].

The capabilities of ML for defect detection over various surfaces with differing levels of complexity, including concrete, asphalt, and masonry, have been assessed with a series of works using classic and deep learning models [8]. Initially, classic machine learning models, such as random forest and multilayer perceptron, were deployed to develop frameworks mainly for crack classification and segmentation. Although the developed frameworks could have significant accuracy in controlled settings, their generalisation was limited as each pixel is treated individually, while cracks are continuous patterns over the surface. Because of this, such frameworks cannot use the spatial connectivity between defect pixels [9,10]. Convolutional neural networks (CNNs) enhanced crack detection by allowing models to use local crack pixel dependencies [11]. By building on convolutional operations, CNNs can extract the local pixel connectivity, thus enhancing the accuracy and applicability in more practical settings. Still, CNNs show a bias towards the texture. CNN-based models [12,13] use texture-based information more than shape-based cues, limiting their performance on unseen data. This contrasts with the human ability to maintain a balance between shape and texture, which can vary depending on the situation [10,14].

Transformers can address the bias towards the texture by incorporating self-attention mechanisms to capture multi-range interdependencies in diverse data types, including images [15]. Self-attention calculates the importance of each element in a sequence by considering its relationship with all other elements, thus allowing the model to focus on different parts of the input as needed [4]. Transformers have demonstrated generalisation capabilities even in scenarios involving manipulated or noisy images [3,4,10]. Nonetheless, the self-attention mechanism of Transformers requires computing the pairwise attention scores between all patches, which leads to a significant computational cost with the long flattened patch embedding length [16]. This computational cost of the self-attention module poses deployment challenges on edge devices with limited processing power and memory, such as drones and mobile phones, thus limiting the development of efficient and effective crack detection solutions.

To address the computational challenges posed by the Transformer's self-attention mechanism, several architectures have been developed to decrease the computational cost. The linear Transformer reduces the complexity of self-attention from quadratic to linear by approximating the SoftMax function and using the kernel method but with the trade-off of performance [17]. Considering the low-order complexity of recurrent neural networks (RNNs) [18], the Structured State Space for Sequences (S4) model [19] integrated the benefits of the Hippo framework [20] and RNNs and then fused linear attention [17] form the Hungry Hungry Hippos (H3) model [21] for more enhanced efficiency and effectiveness of sequence modelling. In addition, similar to the H3 model, the retentive network (RetNet) reduces the inner S4 layer with a state dimension equal to one, which can be regarded as a specific case of a linear state space model [22]. The Mamba [23] architecture represents a recent further evolution in sequence modelling, enabling the integration and enhancement of the strengths of previous S4/H3 models by introducing varying time steps to optimise performance. The new-generation architecture of Mamba has enabled the development of more efficient and effective visual models [24].

Efficient vision models are essential for crack detection, especially for real-time processing, which necessitates the integration of generalisable models on devices with limited computational capabilities. However, while recent developments in the literature can achieve human-level adaptive generalisation across several material surface cracks, they are often computationally expensive [4]. Apart from model capabilities, another essential factor contributing to the generalisation and high accuracy of models is higher input image resolutions [25], as they contain more contextual information and more accurate semantic representations. Yet, higher resolution requires greater computational power. These considerations necessitate the development of efficient models so that the limited computational power of edge devices can be utilised for higher-resolution processing [26]. Given the optimised performance of the state-space model for large inputs compared to previous architectures, we propose a Mamba-based framework for crack segmentation and demonstrate that, with a smaller number of parameters and floating-point operations, the proposed framework can achieve high segmentation accuracy. Our assumption is that a VMamba-based crack segmentation framework can achieve the same stable and advanced performance as Transformer-based models across different material surface cracks yet with significantly fewer parameters and reduced computational power requirements.

The structure of this paper is outlined as follows: Section 2 presents the implementation of the performance study and discusses the differences among CNN, Transformer, and Mamba. Section 3 details the experimental results with a focus on the analysis of the high efficiency and accuracy of the Mamba-based network. Finally, Section 4 draws the main conclusions and suggests future developments.

**2. Study design**

This section illustrates and discusses how the Mamba architecture operates – particularly the 2D-Select-Scan module, which is the core of the VMamba framework proposed in this paper. Subsequently, the computational complexity of CNN, Transformer, and Mamba-based architectures are analysed, highlighting the great potential of Mamba for an efficient network design. The architectures of CNN and Transformer are introduced in Appendix A.1. A baseline VMamba-based encoder-decoder network is deployed for crack segmentation. The evaluation metrics, datasets and training settings are presented in the next parts of this section. Our implemented framework, which consists of two primary stages (training and test), is outlined in Fig. 1. The model is first trained with the help of the Dice loss function (see Fig. 1-a) and pre-set hyperparameters and then evaluated in three commonly utilised surface crack datasets (concrete, asphalt, and masonry) with different background textures (see Fig. 1-b).

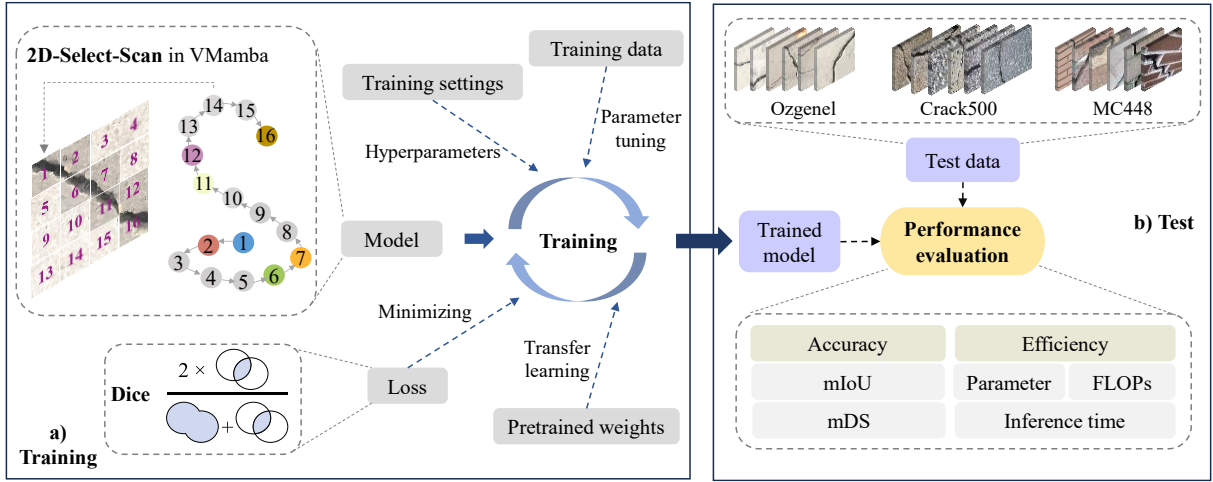

Fig. 1 A diagram of the implemented framework.

## 2.1 The architecture of Mamba

*2.1.1 Vision Mamba (VMamba)*

Based on the recent Select-Scan Structured State Space for Sequences (S6) model, Mamba has been developed as a more efficient choice than CNNs and Transformers for natural language processing [23]. Vision Mamba (VMamba), with a 2D-Select-Scan (SS2D) module, has been introduced for vision tasks, showing high accuracy with a significantly enhanced efficiency [24]. As shown in Fig. 2, the SS2D module starts with a scan expansion, including four routes of top-left to bottom-right, top-right to bottom-left, bottom-left to top-right, and bottom-right to top-left scanning, to obtain four sets of image patches. After patch embedding, each patch $x$ is transformed into a sequence (sequence length: $L$, dimension: $d$) before going through the S6 block:

$$h_k = \overline{A}\, h_{k-1} + \overline{B} x_k , \qquad (6)$$

$$y_k = C h_k + D x_k , \qquad (7)$$

$$\overline{A} = e^{\Delta A}, \qquad (8)$$

$$\overline{B} = (\Delta A)^{-1}(e^{\Delta A} - I) \cdot \Delta B , \qquad (9)$$

where the matrices $B \in \mathbb{R}^{L \times H}$ and $C \in \mathbb{R}^{L \times H}$ with a hidden state size $H$, varying time step $\triangle \in \mathbb{R}^{L \times d}$, are obtained from the linear projection of the input sequence $x \in \mathbb{R}^{L \times d}$. The matrix $A \in \mathbb{R}^{d \times H}$ for hidden state space and $D \in \mathbb{R}^{L \times L}$ for the skip connection consist of trainable parameters that are updated during backpropagation. The outputs of the S6 block are then transformed into the original size of the input through the scan merging operation.

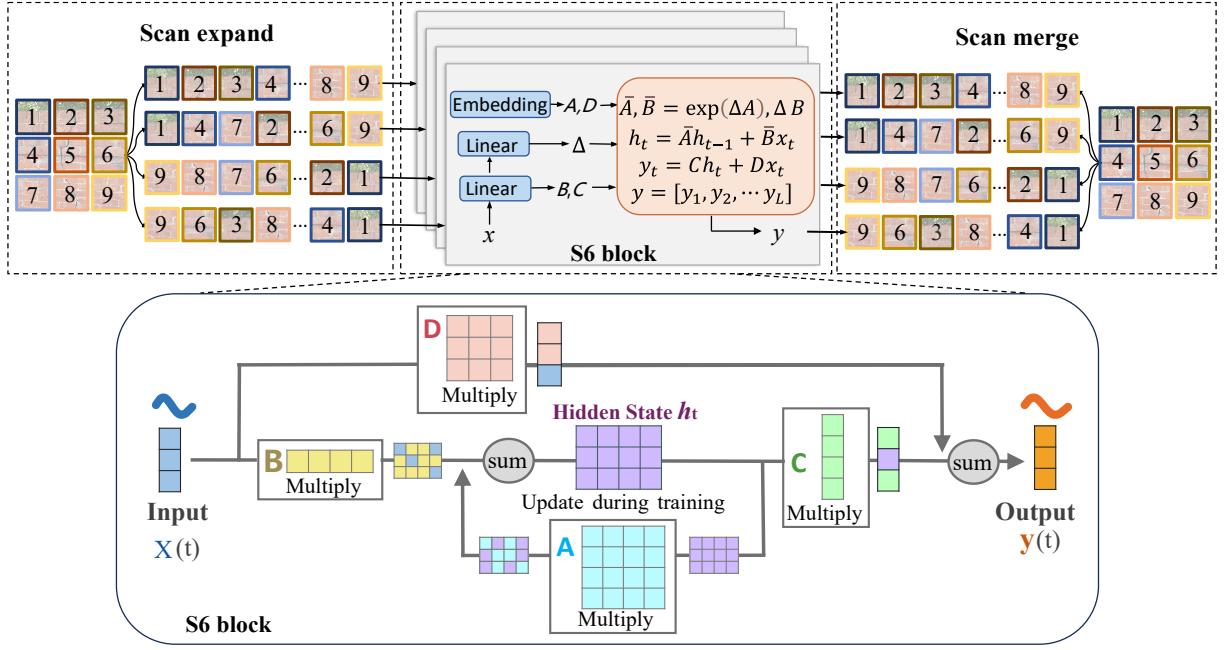

Fig. 2 Illustration of the 2D-Select-Scan (SS2D) in VMamba.

The S6 block can be transformed to operate in another format of Q, K, and V matrices [24]. For the sequences $X := [x_a; \ldots; x_b]$ with length $L$, the following variables are defined as:

$$V := [V_1; \ldots; V_L] \in \mathbb{R}^{L \times d_v}, \text{ where } V_i = x_{a+i-1}\Delta_{a+i-1} ; \tag{10}$$

$$K := [K_1; \ldots; K_L] \in \mathbb{R}^{L \times d_k}, \text{ where } K_i = B_{a+i-1} ; \tag{11}$$

$$Q := [Q_1; \ldots; Q_L] \in \mathbb{R}^{L \times d_k}, \text{ where } Q_i = C_{a+i-1} ; \tag{12}$$

$$w := [w_1; \ldots; w_L] \in \mathbb{R}^{L \times d_k \times d_v}, \text{ where } w_i = \prod_{j=1}^{i} e^{A\Delta_{a-1+j}} ; \tag{13}$$

$$H := [h_a; \ldots; h_b] \in \mathbb{R}^{L \times d_k \times d_v} ; \tag{14}$$

$$Y := [y_a; \ldots; y_b] \in \mathbb{R}^{L \times d_v} ; \tag{15}$$

based on the definitions, the time-varying hidden state can be written as:

$$h_b = w_L \odot h_a + \sum_{i=1}^{L} \frac{w_L}{w_i} \odot (K_i^T V_i) , \tag{16}$$

where $\odot$ is the element-wise production and the division of $w_L$ and $w_i$ is also element-wise. The output of the S6 block can be expressed as:

$$y_b = Q_L(w_L \odot h_a) + Q_L \sum_{i=1}^{L} \frac{w_L}{w_i} \odot (K_i^T V_i) . \tag{17}$$

Thus, the $j^{th}$ slice along the dimension $d_v$ of output $Y^{(j)}$ is:

$$Y^{(j)} = (Q_L \odot w^{(j)})h_a^{(j)} + \left[(Q_L \odot w^{(j)})\left(\frac{K}{w^{(j)}}\right)^T \odot M\right] V^{(j)} , \tag{18}$$

where $M$ presents the temporal mask matrix with the lower triangular parts set to 1 and elsewhere 0. As shown in Eq. (18), the multiplication of Q, K, and V matrices without SoftMax function is similar to the self-attention module of Vision Transformers in Eq. (3) and is regarded as the generalised form of Gated Linear Attention [24,27]. This linear attention in VMamba contributes to its computational efficiency and considerable performance enhancement in multiple vision assignment types.

*2.1.2 Computational complexity comparison*

Due to the discrepancy among CNNs, Transformers, and Mamba, the computational complexity of image processing by each of these architectures is different. As presented in Fig. 3, the input and output images are assumed to have the same size and are equally flattened into several patches, with each patch embedding size *L*. To keep the same size of input and output, the convolutional operation in CNNs can be regarded as *L* times dot-productions of two flattened vectors with size *k*, where *k* is the convolutional kernel size, thus resulting in its $O(kL)$ computational complexity [16]. In the self-attention module of Transformers, after concatenating Queries (Q), Keys (K), and Values (V) matrices, each embedded image patch needs to dot-product all patch embeddings to get the attention score, and the operation of SoftMax($QK^T$) makes the computational complexity become $O(L^2)$ [24]. Unlike Transformers, the S6 block in Mamba – see Eq. (18) – could remove the SoftMax operation and first conduct the operation of $K^T$ and V instead of Q and $K^T$, thus resulting in a linear computational complexity of $O(L)$ [24]. The lower-order computational complexity of the core part contributes to the significant efficiency of Mamba with much lower floating-point operations (FLOPs) than the other two architectures, which will be discussed later.

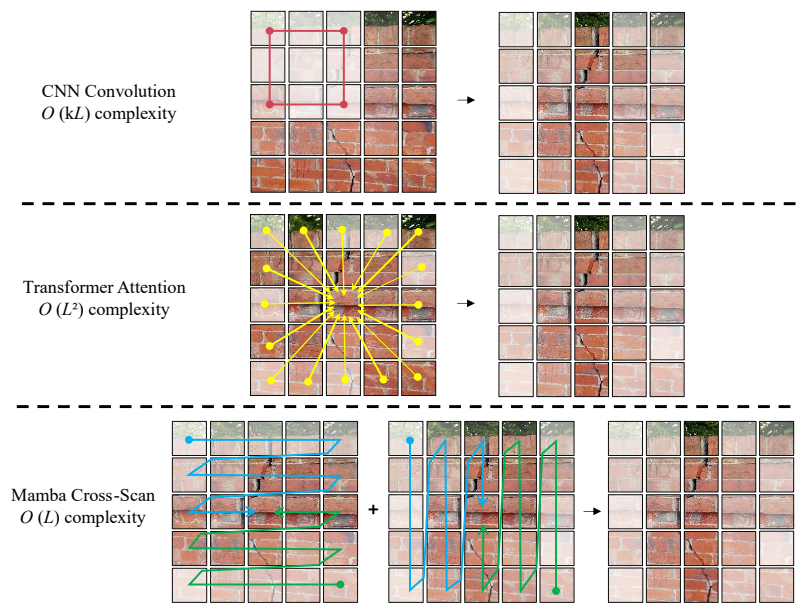

Fig. 3 – Comparison of computational complexity.

## 2.2 Vision Mamba based encoder-decoder network

In this section, the VMamba-based baseline encoder-decoder network (VM-UNet [28]) is introduced for the crack segmentation task. As illustrated in Fig. 4, the patch embedding layer of VM-UNet first partitions the input image into non-overlapping patches with size 4 × 4, then transforms the image dimensions into C. The embedded image is then fed into the encoder, which consists of four stages of feature extraction, with a patch merging down-sampling module [29] employed for the initial three stages to decrease the size of the input feature maps while augmenting the channels. Likewise, the decoder part is structured into four stages, with a patch expanding up-sampling module [30] applied to reduce the channel number while enlarging the feature map size.

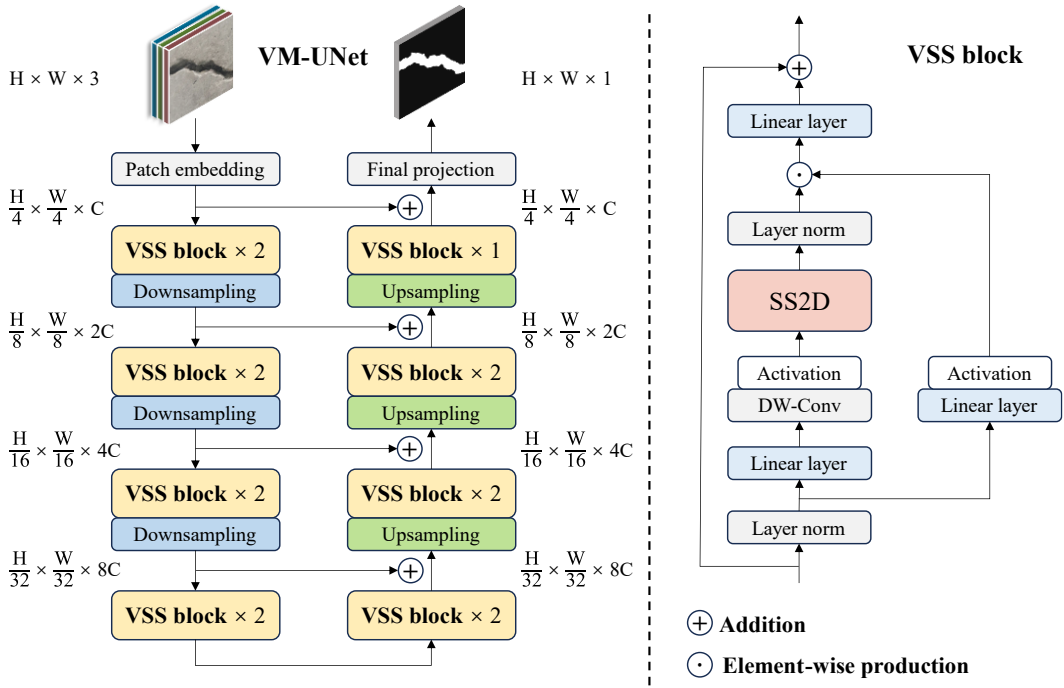

Fig. 4 Overview of the VM-UNet [28] architecture.

The Visual State Space (VSS) block [24] in the encoder and decoder serves as the core component of VM-UNet. After layer normalisation, the input feature map undergoes division into two pathways. In the initial path, the input feature map proceeds through a linear layer and then the Sigmoid Linear Unit (SiLU) [31] activation function. Meanwhile, in the second path, the input feature map undergoes processing via a linear layer, depth-wise separable convolution, and the SiLU activation function, followed by the SS2D (See Fig. 2) module and layer normalisation for additional feature extraction. Afterwards, the element-wise production is operated to merge the outputs of these two pathways, followed by a linear layer and the combination with the residual connection to shape the final output.

## 2.3 Evaluation metrics

For the crack segmentation task, which aims to classify the crack and non-crack pixels, mean Dice score (mDS) and mean intersection over union (mIoU) score are commonly utilised due to their balanced evaluation for segmentation accuracy [10]. Thus, these two metrics are employed to assess the model performance in this study, which can be formulated as:

$$DS = \frac{2|P \cap T|}{|P|+|T|}, \quad (10)$$

$$IoU = \frac{|P \cap T|}{|P \cup T|}, \quad (11)$$

herein, $P$ is the pixel map of the model output, and $T$ is the ground truth pixel map.

IoU measures the overlap between mask pixels and segmentation map sets by dividing their intersection area by union, whereas DS measures overlap by dividing twice the intersection area by the total area of both sets. Both metrics range from 0 to 1, where 1 indicates perfect overlap, and 0 indicates no overlap. IoU penalises poor detections more than DS, especially when overlaps are small. However, as overlaps increase, the difference between IoU and DS decreases. Therefore, mIoU is more sensitive to poor predictions, while mDS shows average performance more accurately. Since these two metrics do not depend on the designated threshold for the classification of crack pixels and assess the overlap of the entire region, including its boundary, they offer a more impartial and reasonable evaluation of segmentation performance [4].

## 2.4 Datasets

In this study, three public crack datasets, with concrete, concrete and asphalt mixed, and masonry backgrounds, are utilised to assess the performance of VMamba-based encoder-decoder network (VM-UNet). Detailed information on the selected datasets is introduced below.

- Crack500 [32]: this dataset depicts cracks on asphalt and concrete pavement surfaces, includes 1896 training images and 1124 testing images. The original images were collected by cell phone cameras on the main campus of Temple University and split by the dataset developers, who retained only the split images (640 × 360) containing crack pixels. To satisfy the input of models analysed in this study, each image and its corresponding annotation were resized to 448 × 448 pixels.

- Ozgenel [33]: this dataset consists of 458 high-resolution concrete crack images (4032 × 3024 pixels) with corresponding annotation masks collected from various Middle East Technical University buildings. In this study, the images were split into patches with a resolution of 448 × 448 pixels to prepare the dataset for evaluation. Finally, the training and test set consist of 1800 and 454 images with their masks, respectively.

- MC448 [3]: this dataset is sourced from 45 copyright-free images, 153 manually captured full-resolution photos, and 15 artificially generated images. After splitting, three thousand images and the corresponding masks were allocated for the model training, and the remaining 351 images and their masks were assigned for testing. This dataset is challenging for crack segmentation because it includes non-crack images and has textures on the masonry surface that resemble cracks, making it harder to differentiate between actual cracks and background textures.

Overall, the specifications of these datasets are presented in Table 1 in terms of the input image size and the image number of training and test sets.

Table 1 Details of three datasets

| Dataset | Input image size | Total number | Train set | Test set |
| --- | --- | --- | --- | --- |
| Crack500 | 448 × 448 | 3020 | 1896 | 1124 |
| Ozgenel | 448 × 448 | 2254 | 1800 | 454 |
| MC448 | 448 × 448 | 3351 | 3000 | 351 |

## 2.5 Training configuration

The models in this study were constructed by Python employing the PyTorch [34] open-source framework. The training and evaluation of models were carried out on the Ubuntu 22.04 LTS system, utilising a workstation equipped with an AMD Ryzen 9 5900X 12-Core CPU, an NVIDIA GeForce RTX 3090 Ti GPU, and 32 gigabytes of RAM. All the models in this study underwent training, validation, and testing within the same framework to ensure a fair comparison. The training settings included a learning rate of $5 \times 10^{-5}$ and a batch size of 2 images, running for 80 epochs with the AdamW optimiser and Dice loss function referred to in Eq. (10). The training set was initially shuffled randomly followed by random image augmentation techniques, such as colour-jittering and horizontal flipping, to prevent over-fitting during the training process. Additionally, to facilitate faster training and enhance the final model performance, the model backbones were pre-trained on the ImageNet [35] dataset and the trained weights were utilised for model initialisation.

## 3. Results and discussions

This section first implements the accuracy comparison for representative CNN, Transformer, and Mamba-based models, thus validating the comparable performance of Mamba-based encoder-decoder network (VM-UNet). An efficiency comparison is then conducted to highlight the advantage of the low-order complexity of Mamba.

## 3.1 Accuracy comparison

Considering the literature advancements achieved in crack segmentation with CNN and Transformer-based models, a comprehensive study is conducted with the following representative architectures as the established benchmarks in the field:

- CNN-based: UNet [36] and LinkNet [37] with the EfficientNet [38] backbone (UNet-EB7 & LinkNet-EB7);
- Transformer-based: SwinUNet [30], and SegFormer-B5 [39];
- CNN-Transformer hybrid-designed: TransUNet [40] and DTrC-Net [41];
- Efficient self-attention designed: PoolingCrack [4] with advanced performances in various crack datasets.

The results of these models are studied and compared – see Table 2 – to a Mamba-based baseline network (VM-UNet) with a simple design to reveal the capabilities of the Mamba architecture. As presented in Table 2, Mamba-based VM-UNet exhibits comparable mDS and mIoU, with up to 0.5% lower mDS and 0.3% - 0.4% lower mIoU than the best ones in the three datasets while having 15.6% - 74.5% fewer training parameters than all other networks. Except for SegFormer-B5 and PoolingCrack, VM-UNet achieves 0.1% - 6.2% higher mDS and up to 6.8% higher mIoU than the other architectures. Although SegFormer-B5 outperforms VM-UNet in the Crack500 dataset, it exhibits up to 0.8 % lower mDS and mIoU in the Ozgenel and MC448 datasets, with around three times more training parameters. Although PoolingCrack shows overall higher accuracy than VM-UNet, it has 15.6% more parameters, requires more floating-point operations, and has a longer inference time, which is discussed in the next section in detail. In summary, the Mamba-based network demonstrates significant potential for effective crack segmentation, particularly due to its highly efficient computational performance.

Fig. 5 illustrates the qualitative analysis results of the different network performances over three datasets. As shown in Fig. 5-a, VM-UNet outputs clear and continuous crack maps in the Crack500 dataset without creating noises in most cases. Although SegFormer-B5 exhibits the best average accuracy (quantitative results) in this dataset, it sometimes cannot portray the crack profile consistently, e.g., the crack prediction maps of SegFormer-B5 in the third and fifth rows. As for the produced crack maps of other networks, they miss some crack pixels in the second-row case or mistakenly classify some background noises as cracks in the third-row case, thus having an unstable performance for the crack pixel classification. VM-UNet, which could achieve the highest mDS in the Ozgenel dataset – Table 2, shows enhanced segmentation results compared to other networks – Fig. 5-b. It presents robustness without being disturbed by surface complexities (red stripe and scratch background texture) in the third and last rows while accurately producing the complete crack map in the fourth-row case with challenging crack pixels.

Table 2 Accuracy comparison of different networks in three datasets (bold values are the best)

| Model | Parameters(M) | Crack500 | | Ozgenel | | MC448 | |
|---|---|---|---|---|---|---|---|
| | | mDS (%) | mIoU(%) | mDS (%) | mIoU(%) | mDS (%) | mIoU(%) |
| UNet-EB7 | 67 | 69.9 | 55.7 | 84.1 | 77.3 | 72.9 | 62.1 |
| LinkNet-EB7 | 64 | 69.9 | 55.6 | 84.6 | 78.1 | 72.0 | 60.9 |
| TransUNet | 106 | 70.2 | 56.0 | 85.3 | 79.2 | 75.3 | 64.7 |
| SwinUNet | 42 | 68.1 | 53.3 | 83.3 | 76.1 | 70.2 | 58.8 |
| SegFormer-B5 | 85 | **70.7** | **56.5** | 84.9 | 78.6 | 75.6 | 64.7 |
| DTrC-Net | 42 | 67.5 | 53.3 | 84.7 | 78.3 | 69.5 | 58.3 |
| PoolingCrack | 32 | 70.6 | 56.4 | **85.7** | **79.7** | **76.2** | **65.5** |
| VM-UNet | **27** | 70.3 | 56.0 | **85.7** | 79.4 | 75.7 | 65.1 |

In addition, according to the output crack masks in the MC448 dataset in Fig. 5-c, although the crack map produced by the VM-UNet for the last case has slight noises, it creates relatively complete crack patterns compared with other network outputs in the other cases. Despite the challenges associated with the MC448 dataset, a masonry crack dataset with many crack-like non-crack patterns, the Mamba-based VM-UNet achieved high accuracy.

Because of the selective mechanism of Mamba [23], the Mamba-based model will automatically select or ignore divided image patches depending on whether these patches include cracks, which is similar to the behaviour of attention scores in Transformers. As a result, the Mamba-based model could have better accuracy than CNNs and similar precision to Transformers. Yet, the low computational complexity of the selective mechanism in Mamba significantly contributes to an enhanced efficiency relative to the Transformer-based models. The corresponding comparison in terms of efficiency is discussed in detail in the next section.

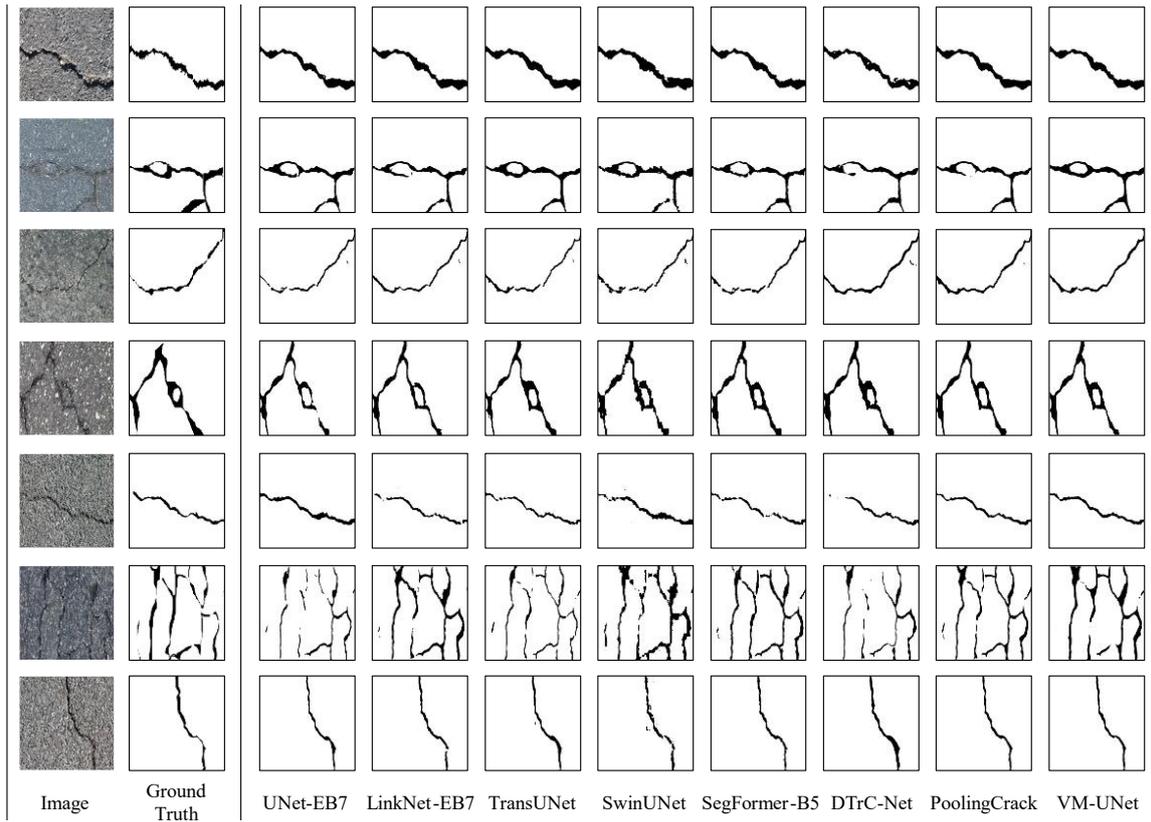

(a) Crack500 (concrete & asphalt)

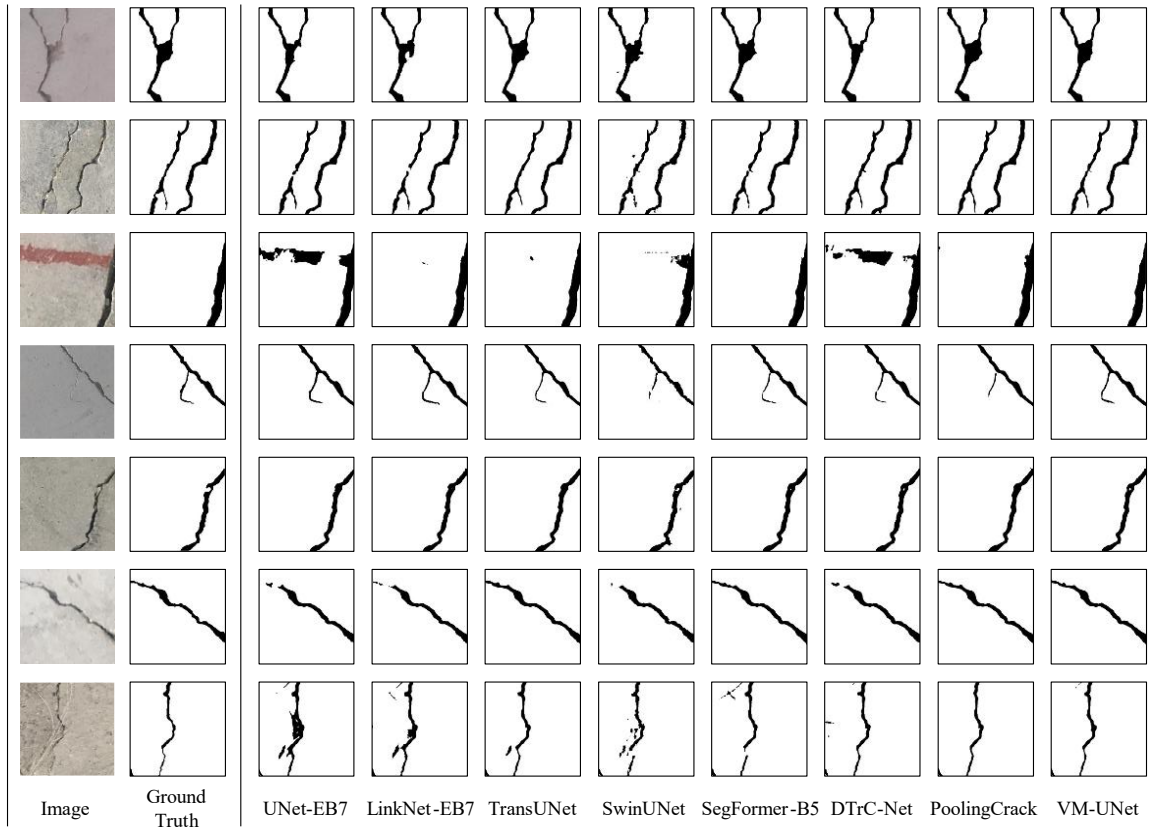

(b) Ozgenel (concrete)

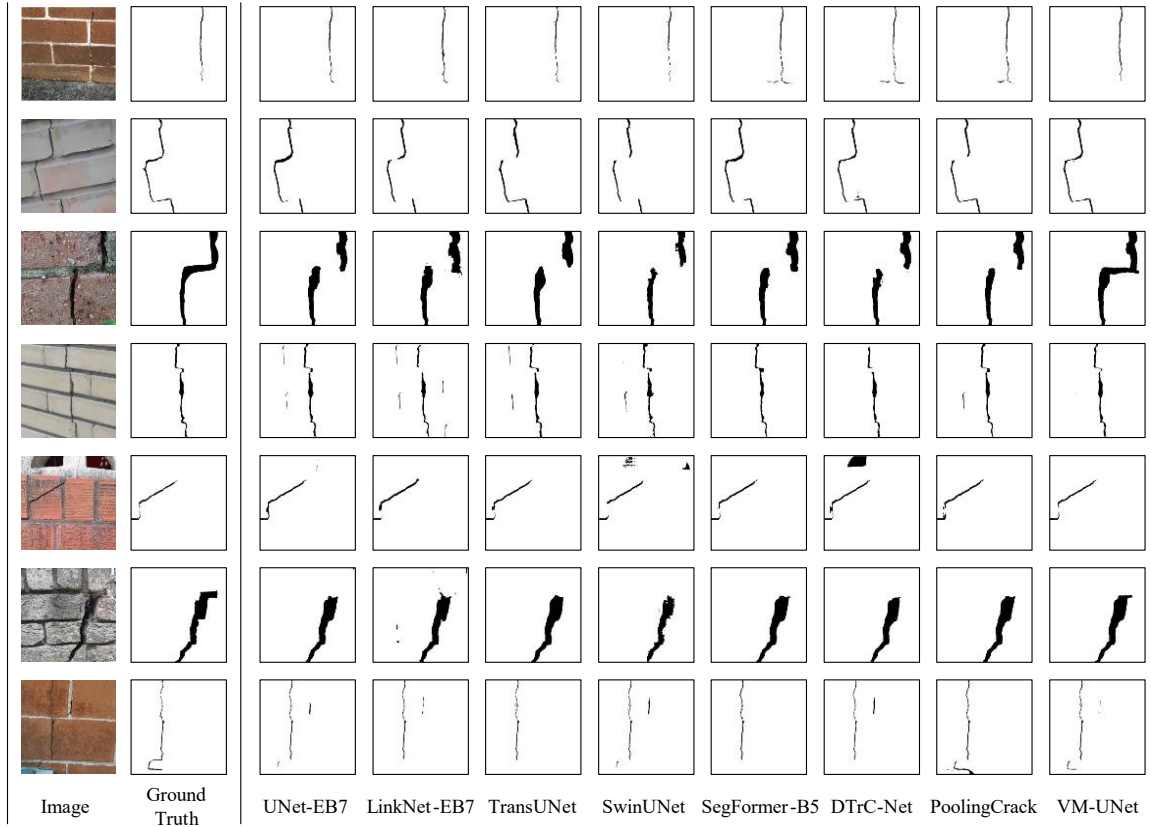

(c) MC448 (masonry)

Fig. 5 Qualitative analysis of the segmentation results for different networks in three datasets: (a) Crack 500, (b) Ozgenel, and (c) MC448.

### 3.2 Efficiency comparison

Three factors, including model parameters (which quantify the complexity of the model), floating-point operations (which measure the computational workload), and inference time (which indicates the time taken for the model to make predictions), are selected as evaluation metrics for the efficiency comparison. The results are presented in Fig. 6.

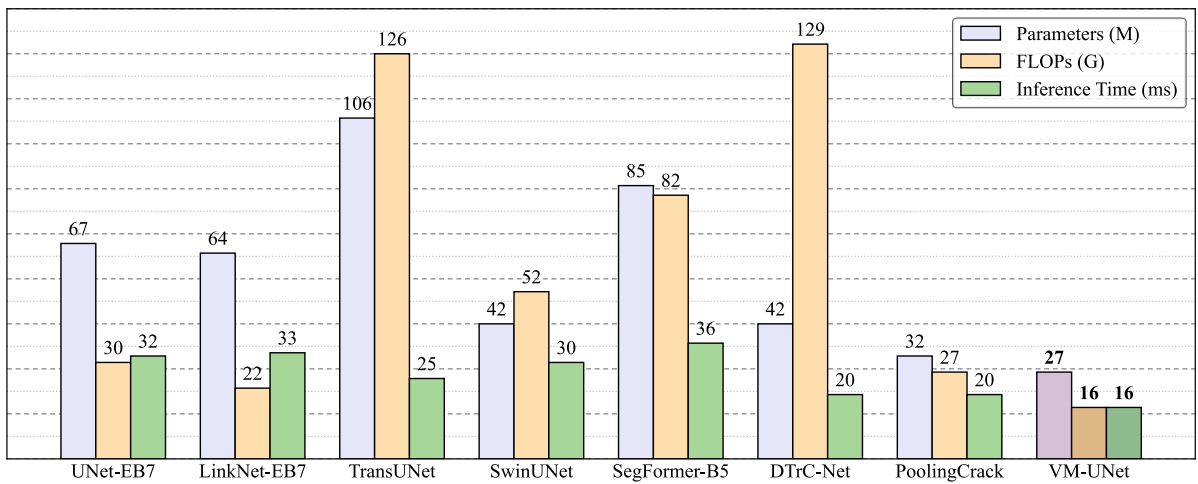

Fig. 6 Efficiency comparison for processing one image with 448 × 448 resolution.

As illustrated in the Fig. 7, the Mamba-based model comprises 27 million training parameters and operates with 16 G FLOPs for each image at a resolution of 448 × 448, achieving an inference time of 16 ms. It has 15.6% - 74.5% fewer training parameters, 27.3% - 87.6% fewer FLOPs, and 20% - 55.6% less inference time than other representative models. Due to the low-order complexity of the Mamba-based architecture, VM-UNet achieves the best efficiency among the similar or even larger scale representative models while showing the same accuracy performance as discussed in the previous section. Another computational cost comparison with different image size inputs, shown in Fig. 7, can better reveal the efficiency of the Mamba-based model; the figure shows the number of FLOPs versus the size of the input image. TransUNet and DTrC-Net (designed with a hybrid of CNN and Transformer) have the top two highest FLOPs as the input image resolution increases. The Transformer-based SwinUNet and SegFormer-B5 present the second-tier highest FLOPs, and the CNN-based UNet-EB7 and LinkNet-EB7 are close to the efficiently designed Transformer-based PoolingCrack in the third tier of highest FLOPs. Mamba-based VM-UNet shows the lowest FLOPs increase among all the models, while it could achieve similar or higher accuracy than the other models – see section 3.1.

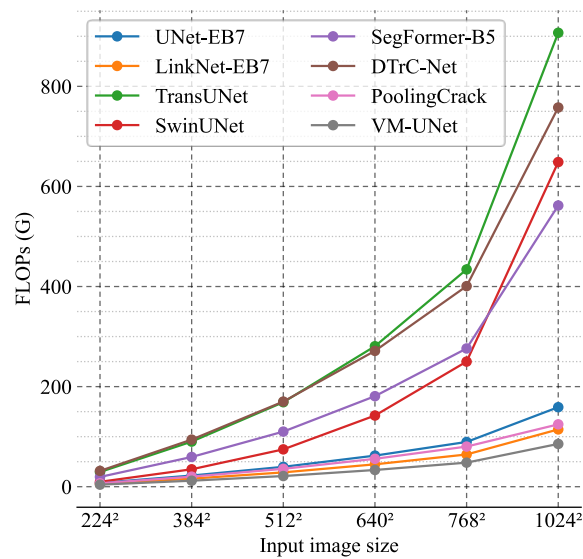

Fig. 7 Computational costs with different image resolution inputs; different line slopes show the discrepancy among the intrinsic complexity of these architectures.

As mentioned in Section 2.1.2, the attention mechanism in Transformer, the convolutional operation in CNN, and the cross-scan in Mamba have the computational complexity of $O(L^2)$, $O(kL)$, and $O(L)$, respectively. Due to the two branches (CNN and Transformer) of image processing, CNN-Transformer hybrid-designed networks exhibit the highest computational costs. Because of the different computational complexity of different cores, the pure Transformer, CNN, and Mamba-based models show in-order decreased computational costs.

Image resolution can be very important when running deep learning models on edge devices and for real-time monitoring, as the capacity of the equipment could be very limited. Representative CNN or Transformer-based models may have few parameters and low FLOPs under the low-resolution image input; nonetheless, the computational cost under high-resolution image input differentiates Mamba from them. For example, DTrC-Net with 42 M parameters has similar FLOPs as VM-UNet with 27 M parameters under 224×224 image input but with up to 9 times the FLOPs of VM-UNet as the input image resolution increases. Similarly, some lightweight Transformer-based models may have fewer parameters and lower FLOPs than Mamba-based models under the low-resolution image input, yet their computational costs would overtake Mamba-based models as the image resolution is enhanced. The intrinsic shortcoming of computational complexity for these traditional CNN or Transformer-based model designs would hinder their further development for future high-resolution crack detection. Based on the above analysis, the Mamba-based model demonstrates potential for efficient and effective crack segmentation.

## 4. Conclusion

Focusing on local information weakens the generalisation of CNN-based models, and the quadratic complexity of the global self-attention hinders the ability of Transformer-based models to process long-sequence inputs. Stemming from these issues, this study proposes a VMamba-based framework for crack segmentation, with low-order complexity of processing crack images and varying time steps to capture crack or ignore non-crack information and improve the generalisation. A VMamba-based encoder-decoder network was used to develop the framework, and its performance was compared with representative CNN-based (UNet-EB7 & LinkNet-EB7), Transformer-based (SwinUNet, SegFormer-B5, and PoolingCrack), and hybrid models (TransUNet & DTrC-Net).

The encoder-decoder network equipped with VMamba (VM-UNet) could achieve up to 4.2% higher, a slightly lower (0.5%), and 0.1% to 6.8% higher mDS and mIoU than representative CNN, Transformer, and CNN-Transformer hybrid-based models while maintaining 15.6% - 74.5% fewer parameters and 20% - 55.6% less inference time. Furthermore, a computational cost comparison was conducted with the increasing network input image sizes. The results showed that the CNN-Transformer hybrid-designed models have the highest FLOPs due to integrating two-branch operations (namely convolution and self-attention), and the Transformer, CNN, and VMamba-based models have the in-order decreasing FLOPs due to the different computational complexity of their cores. As the input image size grows, the VMamba-based model could achieve up to 90.6% lower FLOPs.

The study implemented in this paper could highlight the potential of VMamba for crack detection assignments. The strength of VMamba in processing long-sequence inputs effectively and efficiently could contribute to the development of high-resolution crack detection expected in the future. Given that VMamba is currently in the early development stage, subsequent studies could explore the advanced network design to further improve its overall performance for defect segmentation tasks.

**Appendix A**

**A.1 Convolutional Neural Networks and Transformers**

*A.1.1 Convolutional Neural Networks (CNNs)*

As the classic deep learning network in crack detection [42-45], CNN architecture typically consists of convolutional, pooling, and fully connected layers – see Fig. 8. The convolutional operation extracts the local representations of input images through the trainable filters, transforming them into essential feature maps – see Eq. (1):

$$O(i, j) = \sum_{k=1}^{m} \sum_{l=1}^{n} I(i+k-1, j+l-1) K(k, l), \tag{1}$$

where $I$ and $K$ stand for the input image with size $M \times N$ and the filter with kernel size $m \times n$, respectively; and $O$ is the feature map with size $(M - m + 1) \times (N - n + 1)$ produced by the convolutional operation. In most CNN-based architectures, convolutional and pooling layers are used to systematically capture multi-level feature representations from the data [11]. The fully connected layers traditionally followed these operations to decide on each input element.

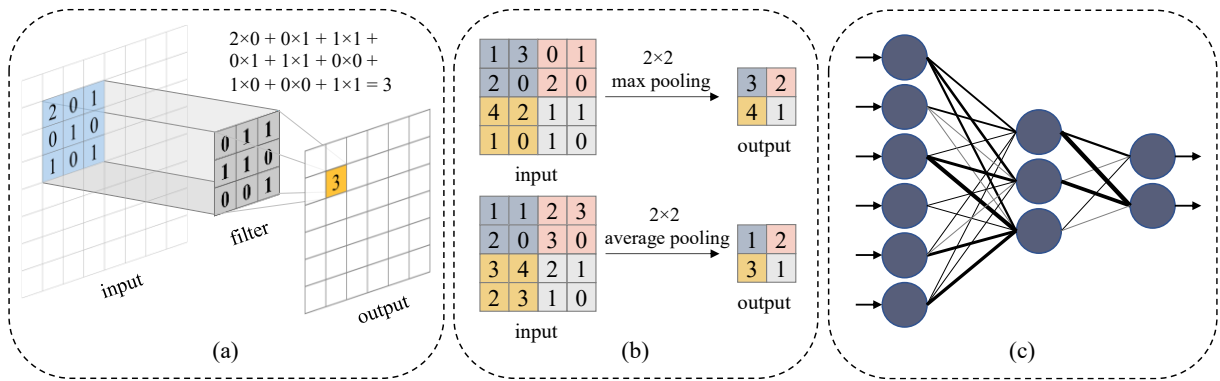

Fig. 8 Diagram of CNN architecture: (a) convolutional operation; (b) max and average pooling; (c) fully connected layers.

CNNs have achieved good results across various vision-used tasks, yet due to the fixed size of the filter kernel and limited receptive field in the convolutional layer, CNNs have the intrinsic attribute of inductive bias, which focuses on the local information and may limit generalisation [10] in settings outside of controlled lab environments. Additionally, the local dependency of CNNs restricts the ability to capture long-range feature information and global context in the input data, thus resulting in sensitivity to noise or irrelevant details [46].

*A.1.2 Vision Transformers*

Transformers have shown significant performance in crack detection [4], which can be primarily attributed to their global attention mechanism. This mechanism facilitates the extraction of multi-range information from inputs and mitigates the biases in CNNs. Initially introduced in natural language processing, transformers have since been adapted for computer vision tasks. A pioneering example of this adaptation is the Vision Transformer [15], which converts input images into sequences by partitioning them into smaller patches before feeding a multi-head self-attention module to extract global contextual information, which is followed by the multilayer perceptron (MLP) [47] to output the prediction results. The schematic diagram of the Vision Transformer is illustrated in Fig. 9.

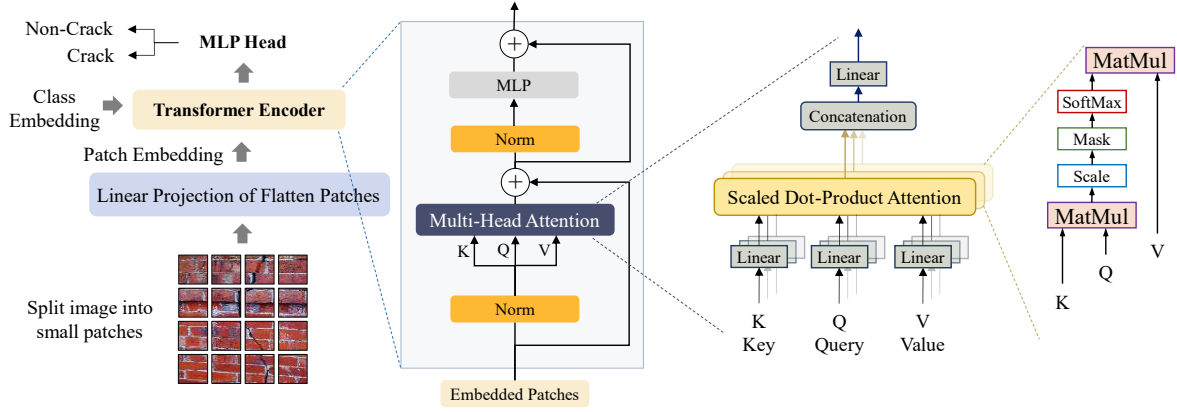

Fig. 9 The schematic diagram of a Vision Transformer.

Assuming the input image is split into $N$ patches, the flattened patches $x_p^i$ ($i = 1, \cdots, N$) with size $p \times p$ are first linearly projected into lower-dimension embeddings as:

$$z_0 = [x_{\text{class}}; x_p^1 E; x_p^2 E; \cdots; x_p^N E] + E_{pos}, \qquad (2)$$

where $x_{\text{class}}$ represents the learnable class embedding, E is the trainable linear projection, and $E_{pos}$ denotes the position embedding. Embedded patch vectors are then multiplied by learnable matrices $W_q$, $W_k$, and $W_v$ to obtain the Key (K), Query (Q), and Value (V) matrices for the self-attention calculation given by:

$$\text{Attention}(Q, K, V) = \text{SoftMax}\left(\frac{Q K^T}{\sqrt{d_k}}\right) V, \qquad (3)$$

where $K^T$ represents the transposed matrix K, $d_k$ denotes the dimension of matrix K. In the Transformer encoder, the multi-head self-attention (MSA) with the linear normalisation (LN) [15] and multilayer perceptron layers are used to extract the feature representations as:

$$z'_l = \text{MSA}(\text{LN}(z_{l-1})) + z_{l-1}, \qquad (4)$$

$$z_l = \text{MLP}(\text{LN}(z'_l)) + z'_l, \qquad (5)$$

where $z_l$ is the encoded feature representation. Although Transformers with self-attention design have shown better generalisation and anti-noise robustness than CNNs [26], the demanding scaled dot-product operations for the calculation of the attention score increase the computational cost, thus impairing model efficiency and restricting the practical deployment on edge devices [4].

**Declaration of Competing Interest**

The authors declare that they have no conflict of interest for the work in this paper.

**Acknowledgement**

The authors acknowledge the Sydney Informatics Hub and the use of the University of Sydney's high performance computing cluster, Artemis. The authors acknowledge the support from the University of Sydney through the Digital Sciences Initiative program.

**References**

[1] M. Azimi, A.D. Eslamlou, G. Pekcan, Data-Driven Structural Health Monitoring and Damage Detection through Deep Learning: State-of-the-Art Review, Sensors 20 (10) (2020), pp. 2778, https://doi.org/10.3390/s20102778.

[2] M. Azimi, G. Pekcan, Structural health monitoring using extremely compressed data through deep learning, Computer-Aided Civil and Infrastructure Engineering 35 (6) (2020), pp. 597-614, https://doi.org/10.1111/mice.12517.

[3] E.A. Shamsabadi, C. Xu, D. Dias-da-Costa, Robust crack detection in masonry structures with Transformers, Measurement 200 (2022), pp. 111590, https://doi.org/10.1016/j.measurement.2022.111590.

[4] Z. Chen, E. Asadi Shamsabadi, S. Jiang, L. Shen, D. Dias-da-Costa, An average pooling designed Transformer for robust crack segmentation, Automation in Construction 162 (2024), pp. 105367, https://www.sciencedirect.com/science/article/pii/S0926580524001031.

[5] Y. Yao, S.T.E. Tung, B. Glisic, Crack detection and characterization techniques—An overview, Structural Control and Health Monitoring 21 (12) (2014), pp. 1387-1413, https://doi.org/10.1002/stc.1655.

[6] F. Elghaish, S. Talebi, E. Abdellatef, S.T. Matarneh, M.R. Hosseini, S. Wu, M. Mayouf, A. Hajirasouli, T.-Q. Nguyen, Developing a new deep learning CNN model to detect and classify highway cracks, Journal of Engineering, Design and Technology 20 (4) (2022), pp. 993-1014, https://doi.org/10.1108/JEDT-04-2021-0192.

[7] F. Elghaish, S.T. Matarneh, S. Talebi, S. Abu-Samra, G. Salimi, C. Rausch, Deep learning for detecting distresses in buildings and pavements: a critical gap analysis, Construction Innovation 22 (3) (2022), pp. 554-579, https://doi.org/10.1108/CI-09-2021-0171.

[8] Y.-A. Hsieh, J. Tsai Yichang, Machine learning for crack detection: review and model performance comparison, Journal of Computing in Civil Engineering 34 (5) (2020), pp. 04020038, https://doi.org/10.1061/(ASCE)CP.1943-5487.0000918.

[9] Y.-J. Cha, W. Choi, O. Büyüköztürk, Deep learning-based crack damage detection using convolutional neural networks, Computer-Aided Civil and Infrastructure Engineering 32 (5) (2017), pp. 361-378, https://doi.org/10.1111/mice.12263.

[10] E.A. Shamsabadi, C. Xu, A.S. Rao, T. Nguyen, T. Ngo, D. Dias-da-Costa, Vision transformer-based autonomous crack detection on asphalt and concrete surfaces, Automation in Construction 140 (2022), pp. 104316, https://doi.org/10.1016/j.autcon.2022.104316.

[11] C.V. Dung, Autonomous concrete crack detection using deep fully convolutional neural network, Automation in Construction 99 (2019), pp. 52-58, https://doi.org/10.1016/j.autcon.2018.11.028.

[12] S. Park, S. Bang, H. Kim, H. Kim, Patch-based crack detection in black box images using convolutional neural networks, Journal of Computing in Civil Engineering 33 (3) (2019), pp. 04019017, https://doi.org/10.1061/(ASCE)CP.1943-5487.0000831.

[13] S. Bang, S. Park, H. Kim, H. Kim, Encoder–decoder network for pixel-level road crack detection in black-box images, Computer-Aided Civil and Infrastructure Engineering 34 (8) (2019), pp. 713-727, https://doi.org/10.1111/mice.12440.

[14] R. Geirhos, P. Rubisch, C. Michaelis, M. Bethge, F.A. Wichmann, W. Brendel, ImageNet-trained CNNs are biased towards texture; increasing shape bias improves accuracy and robustness, arXiv preprint arXiv:1811.12231 (2018), https://arxiv.org/abs/1811.12231.

[15] A. Dosovitskiy, L. Beyer, A. Kolesnikov, D. Weissenborn, X. Zhai, T. Unterthiner, M. Dehghani, M. Minderer, G. Heigold, S. Gelly, An image is worth 16x16 words: Transformers for image recognition at scale, arXiv preprint (2020), https://arxiv.org/abs/2010.11929.

[16] R. Mohammadi Farsani, E. Pazouki, A transformer self-attention model for time series forecasting, Journal of Electrical and Computer Engineering Innovations 9 (1) (2020), pp. 1-10, https://doi.org/10.22061/jecei.2020.7426.391.

[17] A. Katharopoulos, A. Vyas, N. Pappas, F. Fleuret, Transformers are rnns: Fast autoregressive transformers with linear attention, International Conference on Machine Learning, (2020), pp. 5156-5165, https://proceedings.mlr.press/v119/katharopoulos20a.html.

[18] M. Schuster, K.K. Paliwal, Bidirectional recurrent neural networks, IEEE Transactions on Signal Processing 45 (11) (1997), pp. 2673-2681, https://ieeexplore.ieee.org/abstract/document/650093.

[19] A. Gu, K. Goel, C. Ré, Efficiently modeling long sequences with structured state spaces, arXiv preprint arXiv:2111.00396 (2021), https://arxiv.org/abs/2111.00396.

[20] A. Gu, T. Dao, S. Ermon, A. Rudra, C. Ré, Hippo: Recurrent memory with optimal polynomial projections, Advances in Neural Information Processing Systems 33 (2020), pp. 1474-1487, https://proceedings.neurips.cc/paper_files/paper/2020/hash/102f0bb6efb3a6128a3c750dd16729be-Abstract.html.

[21] D.Y. Fu, T. Dao, K.K. Saab, A.W. Thomas, A. Rudra, C. Ré, Hungry hungry hippos: Towards language modeling with state space models, arXiv preprint arXiv:2212.14052 (2022), https://arxiv.org/abs/2212.14052.

[22] Y. Sun, L. Dong, S. Huang, S. Ma, Y. Xia, J. Xue, J. Wang, F. Wei, Retentive network: A successor to transformer for large language models, arXiv preprint arXiv:2307.08621 (2023), https://arxiv.org/abs/2307.08621.

[23] A. Gu, T. Dao, Mamba: Linear-time sequence modeling with selective state spaces, arXiv preprint arXiv:2312.00752 (2023), https://arxiv.org/abs/2312.00752.

[24] Y. Liu, Y. Tian, Y. Zhao, H. Yu, L. Xie, Y. Wang, Q. Ye, Y. Liu, Vmamba: Visual state space model, arXiv preprint arXiv:2401.10166 (2024), https://arxiv.org/abs/2401.10166.

[25] H. Chu, P.j. Chun, Fine-grained crack segmentation for high-resolution images via a multiscale cascaded network, Computer-Aided Civil and Infrastructure Engineering (2023), https://doi.org/10.1111/mice.13111.

[26] Z. Chen, E.A. Shamsabadi, S. Jiang, L. Shen, D. Dias-da-Costa, Robust feature knowledge distillation for enhanced performance of lightweight crack segmentation models, arXiv preprint arXiv:2404.06258 (2024), https://arxiv.org/abs/2404.06258.

[27] S. Yang, B. Wang, Y. Shen, R. Panda, Y. Kim, Gated linear attention transformers with hardware-efficient training, arXiv preprint arXiv:2312.06635 (2023), https://arxiv.org/abs/2312.06635.

[28] J. Ruan, S. Xiang, Vm-unet: Vision mamba unet for medical image segmentation, arXiv preprint arXiv:2402.02491 (2024), https://arxiv.org/abs/2402.02491.
[29] Z. Liu, H. Hu, Y. Lin, Z. Yao, Z. Xie, Y. Wei, J. Ning, Y. Cao, Z. Zhang, L. Dong, Swin transformer v2: Scaling up capacity and resolution, Proceedings of the IEEE/CVF Conference on Computer Vision and Pattern Recognition (2022), pp. 12009-12019, https://openaccess.thecvf.com/content/CVPR2022/html/Liu_Swin_Transformer_V2_Scaling_Up_Capacity_and_Resolution_CVPR_2022_paper.html.
[30] H. Cao, Y. Wang, J. Chen, D. Jiang, X. Zhang, Q. Tian, M. Wang, Swin-unet: Unet-like pure transformer for medical image segmentation, European Conference on Computer Vision, Springer, (2022), pp. 205-218, https://link.springer.com/chapter/10.1007/978-3-031-25066-8_9.
[31] S. Elfwing, E. Uchibe, K. Doya, Sigmoid-weighted linear units for neural network function approximation in reinforcement learning, Neural networks 107 (2018), pp. 3-11, https://doi.org/10.1016/j.neunet.2017.12.012.
[32] L. Zhang, F. Yang, Y.D. Zhang, Y.J. Zhu, Road crack detection using deep convolutional neural network, 2016 IEEE International Conference on Image Processing (2016), pp. 3708-3712, https://ieeexplore.ieee.org/document/7533052.
[33] Ç.F. Özgenel, Concrete crack segmentation dataset, Mendeley Data 1 (2019), pp. 2019, https://data.mendeley.com/datasets/jwsn7tfbrp/1.
[34] A. Paszke, S. Gross, F. Massa, A. Lerer, J. Bradbury, G. Chanan, T. Killeen, Z. Lin, N. Gimelshein, L. Antiga, Pytorch: an imperative style, high-performance deep learning library, Advances in Neural Information Processing Systems 32 (2019), https://proceedings.neurips.cc/paper_files/paper/2019/hash/bdbca288fee7f92f2bfa9f7012727740-Abstract.html.
[35] J. Deng, W. Dong, R. Socher, L.-J. Li, K. Li, L. Fei-Fei, Imagenet: a large-scale hierarchical image database, 2009 IEEE Conference on Computer Vision and Pattern Recognition, (2009), pp. 248-255, https://ieeexplore.ieee.org/document/5206848.
[36] O. Ronneberger, P. Fischer, T. Brox, U-net: convolutional networks for biomedical image segmentation, Medical Image Computing and Computer-Assisted Intervention–18th International Conference, Munich, Germany Springer, (2015), pp. 234-241, https://link.springer.com/chapter/10.1007/978-3-319-24574-4_28.
[37] A. Chaurasia, E. Culurciello, Linknet: Exploiting encoder representations for efficient semantic segmentation, 2017 IEEE visual communications and image processing (VCIP), IEEE, (2017), pp. 1-4.
[38] M. Tan, Q. Le, Efficientnet: rethinking model scaling for convolutional neural networks, Proceedings of Machine Learning Research, (2019), pp. 6105-6114, http://proceedings.mlr.press/v97/tan19a.html?ref=jina-ai-gmbh.ghost.io.
[39] E. Xie, W. Wang, Z. Yu, A. Anandkumar, J.M. Alvarez, P. Luo, SegFormer: simple and efficient design for semantic segmentation with transformers, Advances in Neural Information Processing Systems 34 (2021), pp. 12077-12090, https://proceedings.neurips.cc/paper/2021/hash/64f1f27bf1b4ec22924fd0acb550c235-Abstract.html.
[40] J. Chen, Y. Lu, Q. Yu, X. Luo, E. Adeli, Y. Wang, L. Lu, A.L. Yuille, Y. Zhou, Transunet: transformers make strong encoders for medical image segmentation, arXiv preprint arXiv:2102.04306 (2021), https://arxiv.org/abs/2102.04306.
[41] C. Xiang, J. Guo, R. Cao, L. Deng, A crack-segmentation algorithm fusing transformers and convolutional neural networks for complex detection scenarios, Automation in Construction 152 (2023), pp. 104894, https://doi.org/10.1016/j.autcon.2023.104894.
[42] R. Ali, J.H. Chuah, M.S.A. Talip, N. Mokhtar, M.A. Shoaib, Automatic pixel-level crack segmentation in images using fully convolutional neural network based on residual blocks

[43] R. Ali, J.H. Chuah, M.S.A. Talip, N. Mokhtar, M.A. Shoaib, Structural crack detection using deep convolutional neural networks, Automation in Construction 133 (2022), pp. 103989, https://www.sciencedirect.com/science/article/pii/S0926580521004404.

[44] Z. Fan, C. Li, Y. Chen, J. Wei, G. Loprencipe, X. Chen, P. Di Mascio, Automatic crack detection on road pavements using encoder-decoder architecture, Materials 13 (13) (2020), pp. 2960, https://doi.org/10.3390/ma13132960.

[45] C. Li, Z. Fan, Y. Chen, H. Lin, L. Moretti, G. Loprencipe, W. Sheng, K.C. Wang, CrackCLF: Automatic Pavement Crack Detection Based on Closed-Loop Feedback, IEEE Transactions on Intelligent Transportation Systems (2023), https://arxiv.org/abs/2311.11815.

[46] D. Zhou, Z. Yu, E. Xie, C. Xiao, A. Anandkumar, J. Feng, J.M. Alvarez, Understanding the robustness in vision transformers, International Conference on Machine Learning, (2022), pp. 27378-27394, https://proceedings.mlr.press/v162/zhou22m.html.

[47] M.W. Gardner, S. Dorling, Artificial neural networks (the multilayer perceptron)—a review of applications in the atmospheric sciences, Atmospheric Environment 32 (14-15) (1998), pp. 2627-2636, https://doi.org/10.1016/S1352-2310(97)00447-0.

and pixel local weights, Engineering Applications of Artificial Intelligence 104 (2021), pp. 104391, https://www.sciencedirect.com/science/article/pii/S0952197621002396.